\documentclass[letterpaper, 10 pt, conference]{ieeeconf}  % Comment this line out if you need a4paper

\IEEEoverridecommandlockouts                              % This command is only needed if 
\usepackage{graphics} % for pdf, bitmapped graphics files
\usepackage{epsfig} % for postscript graphics files
\usepackage{mathptmx} % assumes new font selection scheme installed
\usepackage{times} % assumes new font selection scheme installed
\usepackage{amsmath} % assumes amsmath package installed
\usepackage{amssymb}  % assumes amsmath package installed
\usepackage{comment}
\usepackage{rotating}
\usepackage{xspace}
\usepackage[colorlinks=true,linkcolor=black,anchorcolor=black,citecolor=black,filecolor=black,menucolor=black,runcolor=black,urlcolor=black]{hyperref}
\newcommand{\sL}{\mathcal{L}}

\DeclareMathOperator*{\argmin}{arg\,min}

\newcommand{\reals}{\mathbb{R}}

\makeatletter
\DeclareRobustCommand\onedot{\futurelet\@let@token\@onedot}
\def\@onedot{\ifx\@let@token.\else.\null\fi\xspace}

\def\eg{\emph{e.g}\onedot} 
\def\ie{\emph{i.e}\onedot}

\usepackage{accents}

\usepackage{xcolor}
\usepackage{multirow}

\makeatother
\usepackage{pifont}
\usepackage{tikz}
\usepackage{pgfplotstable}
\definecolor{blue(munsell)}{rgb}{0.0, 0.5, 0.69}

\let\labelindent\relax
\usepackage{enumitem}

\title{\LARGE \bf
Contrastive Learning for Cross-Domain Open World Recognition
}

\author{Francesco Cappio Borlino$^{1}$, Silvia Bucci$^{1}$, and Tatiana Tommasi$^{1}$
\thanks{$^{1}$F. Cappio Borlino, S. Bucci and T. Tommasi are with the DAUIN Department at Politecnico di Torino, Italy. F. Cappio Borlino and T. Tommasi are also affiliated with the Italian Institute of Technology, Italy. 
        {\tt\small \{francesco.cappio, silvia.bucci, tatiana.tommasi\}@polito.it}}%%
}

\begin{document}

\newcommand{\OUR}{COW\xspace}

\maketitle
\thispagestyle{empty}
\pagestyle{empty}

%%%%%%%%%%%%%%%%%%%%%%%%%%%%%%%%%%%%%%%%%%%%%%%%%%%%%%%%%%%%%%%%%%%%%%%%%%%%%%%%
\begin{abstract}

The ability to evolve is fundamental for any valuable autonomous agent whose knowledge cannot remain limited to that injected by the manufacturer. Consider for example a home assistant robot: it should be able to incrementally learn new object categories when requested, but also to recognize the same objects in different environments (rooms) and poses (hand-held/on the floor/above furniture), while rejecting unknown ones. Despite its importance, this scenario has started to raise interest in the robotic community only recently and the related research is still in its infancy, with existing experimental testbeds but no tailored methods.
With this work, we propose the first learning approach that deals with all the previously mentioned challenges at once  by exploiting a single contrastive objective. We show how it learns a feature space perfectly suitable to incrementally include new classes and is able to capture knowledge which generalizes across a variety of visual domains. 
Our method is endowed with a tailored effective stopping criterion for each learning episode and exploits a self-paced thresholding strategy that provides the classifier with a reliable rejection option. Both these novel contributions are based on the observation of the data statistics and do not need manual tuning. 
An extensive experimental analysis confirms the effectiveness of the proposed approach in establishing the new state-of-the-art. The code is available at  \url{https://github.com/FrancescoCappio/Contrastive_Open_World}.
\end{abstract}

%%%%%%%%%%%%%%%%%%%%%%%%%%%%%%%%%%%%%%%%%%%%%%%%%%%%%%%%%%%%%%%%%%%%%%%%%%%%%%%%
\section{INTRODUCTION}

Trustworthy robotic assistants for industrial, home and street environments should be able to recognize multiple objects and detect new unseen categories. Although we are taking as example scenarios that offer different levels of control, all of them share the need for robust visual systems: 
they should generalize to a variety of real-world target conditions (\eg change in viewpoint, camera equipment, illumination, home, weather, country) while maintaining the ability to identify any unknown object and possibly learn its category over time. 
For a practical example let's consider a home assistant robot. It cannot be limited to recognize only a pre-defined set of classes for which it was programmed: it should be able to incrementally learn new objects when its owner handles them and then to recognize those objects when they are naturally arranged in different rooms, without getting confused by others for which it has not received instructions \cite{COSDA_WACV_2021}.
Managing all these tasks at once is extremely challenging. Most of the existing successful deep learning models are applied on simplified problems under \emph{closed-set} and \emph{closed-domain} conditions. 
The former considers a match between the training and test category sets so that the classes available while learning are assumed to be the only ones that could be ever encountered at deployment time. The latter means neglecting situations in which train and test samples have the same semantic content but come from different visual distributions (also known as domains). 

\begin{figure}[tb]
\vspace{-3mm}
\centering
\includegraphics[width=0.43\textwidth]{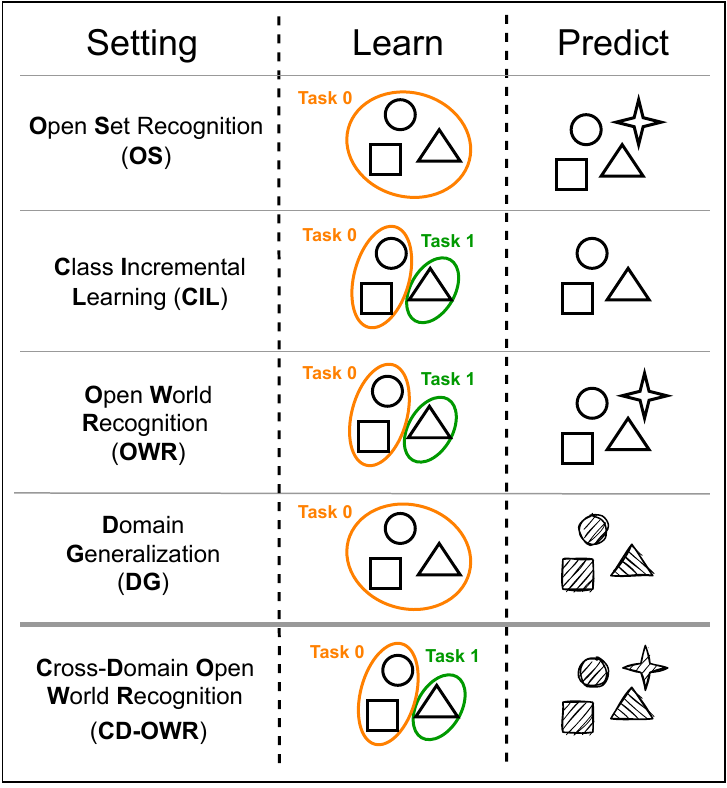}
    \caption{Cross-Domain Open World Recognition (CD-OWR) and other related settings compared with respect to the training and deployment conditions. Each shape indicates a different category. The sketchy style specifies a different data distribution with respect to that of the white empty shapes. The task index higher than 0 refers to the settings where multiple incremental learning steps are needed.}\vspace{-3mm}
    \label{fig:new_teaser}
\vspace{-3mm}
\end{figure}

Several learning frameworks have been defined to push the boundaries of object recognition towards more realistic \emph{open-world} scenarios (see Figure \ref{fig:new_teaser}).
\emph{Open-Set recognition} (OS) deals with the identification of novel classes at test time that were not present in the training phase, while also maintaining the recognition performance for known classes \cite{bendale-boult-cvpr2016,gopenmax}. 
\emph{Class Incremental Learning} (CIL) 
focuses on extending an original model to accommodate novel classes in subsequent incremental tasks \cite{rebuffi-cvpr2017,li2017learning,ahn2021ss}. 
Existing works on \emph{Open World Recognition} (OWR) combine OS and CIL but mainly disregard domain shift conditions \cite{ncm,mancini2019knowledge,fontanel2020boosting}. Indeed, a change in domain between training and test data can create confusion in the identification of the novel categories, and consequently, make their inclusion in the training process even more challenging.  

\emph{Cross-Domain Learning} (CD) aims at improving the performance of a model trained on a labeled source domain when tested on data of a distinct target domain \cite{csurka2017domain,wang2018deep}. 
Several \emph{Domain Generalization} (DG) and \emph{Domain Adaptation} (DA) approaches have been developed for this purpose, with DG methods working without accessing the unlabeled target data at training time. An extension of DG to the open-set scenario was recently presented in \cite{shu2021open}, but the proposed method strongly relies on the availability of multiple data sources.
There have been also some attempts to define a wider \emph{Cross-Domain Open-World Recognition} (CD-OWR) setting with works mainly focusing on benchmarks and na\"{i}ve combinations of existing methods (OWR+DA \cite{COSDA_WACV_2021}, OWR+DG \cite{fontanel_owr_cross_domain}): they have highlighted how difficult the setting is, remaining far from solving it.

\begin{figure*}[t!]
\centering
\includegraphics[width=0.95\textwidth]{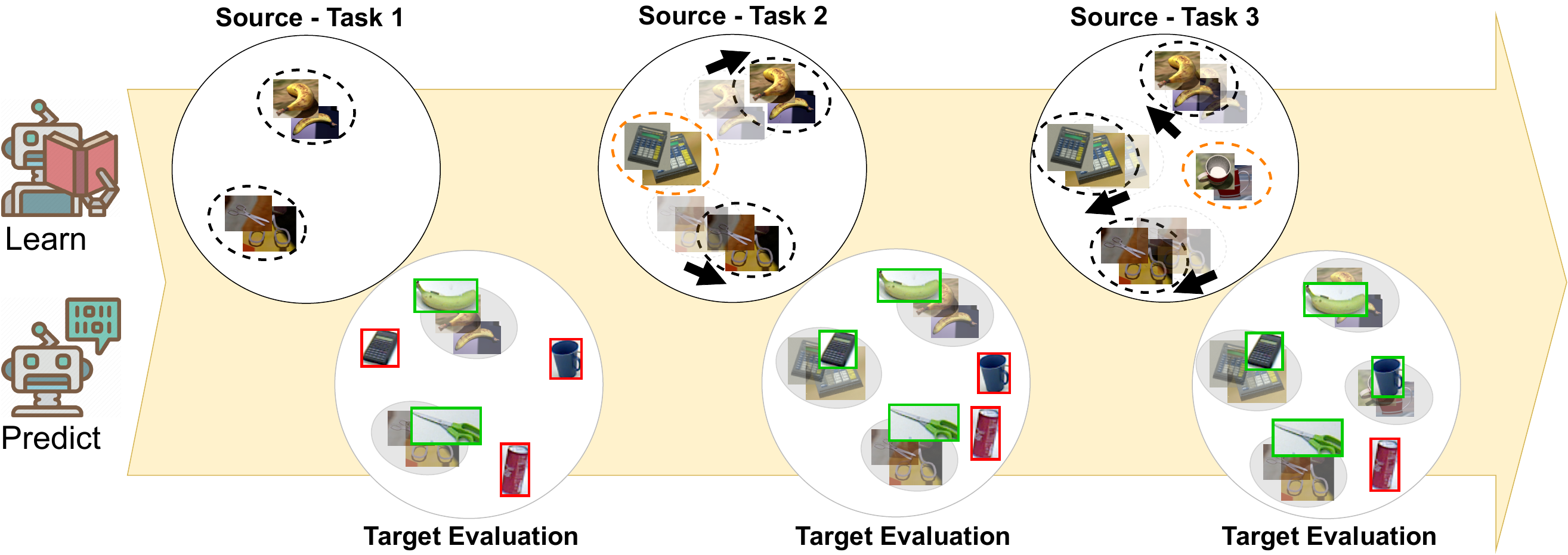}\vspace{-1mm}
    \caption{Overview of \OUR. During training, the classes are incrementally learned at each time step and the old clusters move on the hypersphere to make room for new semantic categories. During the evaluation the target is mapped on the trained hypersphere: the samples far from any existing centroid are marked as unknown (red square), and the samples near enough to a centroid are classified as known with the category of the nearest one (green square).} \vspace{-3mm}
    \label{fig:method}
\end{figure*}

To the best of our knowledge, we propose the first approach that deals with all the challenges of the CD-OWR scenario at once.
We show how a single supervised contrastive objective is suitable for open world recognition while also promoting domain generalization.
Specifically, in the hyperspherical feature space obtained via contrastive learning, samples of the same class tend to cluster together regardless of their domain, while novel categories appear in low-density regions (see Fig. \ref{fig:method}). By considering the Nearest Class Mean \cite{ncm} 
logic which is the basis of many OWR methods, it becomes clear that the described embedding is an ideal environment where a simple rejection rule can be applied on sample-to-prototype distances to identify novel categories. Moreover, our approach does not need class-specific rejection thresholds as the learned feature space pushes all clusters to have similar structures and distances, further simplifying the task. 
Our key contributions are the following:
\begin{itemize}[leftmargin=*]
    \item We propose our \emph{Contrastive Open-World} (COW) approach: it is a straightforward method (see Table \ref{tab:comparison}) able to deal with CD-OWR by simply exploiting the highly structured feature space obtained via contrastive learning.
    \item COW manages incremental class learning by using a tailored stopping criterion at each learning episode. 
    Its implementation provides empirical guarantees on the quality of the model learned after each incremental step.
    \item COW exploits a novel thresholding strategy free from ma\-nual tuning
    which effectively separates known and unknown target data. The decision 
    adapts to the peculiarities of the considered datasets since it is based on the observed sample distribution in the learned feature space. 
    \item A thorough experimental analysis of existing CD-OWR benchmarks confirms the effectiveness of COW which sets the new state-of-the-art.
\end{itemize}

\begin{table}[tb]
\vspace{2mm}
\begin{center}
\caption{Comparison with existing OWR (OS+CIL), DG and CIL approaches. HPs indicate the hyperparameters. }
\label{tab:comparison}
\resizebox{\linewidth}{!}{
{
\renewcommand\arraystretch{1.1}

\begin{tabular}{|c@{~}|c@{~~~}|c@{~~~}|c@{~~~}|c@{~~~}|c@{~~~}|}
\hline
\multirow{2}{*}{Method} & {No. of} & {No. of} &  {Open-Set} & {Domain} & {Class Incremental}   \\
& {Losses } & {HPs} & {Recognition} & {Generalization} & {Learning}\\
\hline
{NNO \cite{bendale2015towards}} &  {1} &	{1} &	{\checkmark } &	{}  & 	{\checkmark }\\
{DeepNNO \cite{mancini2019knowledge}} & 	{2} & 	{1} & {\checkmark}  &	{ }  & 	{\checkmark }\\
{B-DOC \cite{fontanel2020boosting}} & 	{3} & 	{2} & 	{\checkmark} &	{ }  & 	{\checkmark }\\
{SS-IL \cite{ahn2021ss}} & 	2 & 	{0} &{ } &	{ }  & 	{\checkmark }\\
\hline
{RR \cite{loghmani2020unsupervised}} & 	{2} & 	{1} & 	{} &	{\checkmark }  & 	{ }\\
{SC \cite{huang2020self}} & {1} & 	{1} & 	{} &	{\checkmark }  & 	{ }\\
{RSDA \cite{volpi2019addressing}} & 	{2} & 	{1} & 	{} &	{\checkmark }  & { }\\
{SagNet \cite{nam2021reducing}} & {3} & 	{2} & 	{} &	{\checkmark }  & 	{ }\\
\hline
{\textbf{\OUR}} & 	{1} & 	{2} & 	{\checkmark} & {\checkmark}  & 	{\checkmark} \\
\hline
\end{tabular}
}
}
\vspace{-9mm}
\end{center}

\end{table}

\section{RELATED WORK}
\subsection{Open World Recognition}
Hard-coded recognition skills are clearly not enough for autonomous robots that operate in unconstrained environments. \emph{Class Incremental} (or \textit{Continual}) \emph{Learning} provides support with strategies that update pretrained models by including new classes once they are observed. In order to be effective, this should be done with as little access to the old data as possible, while also avoiding to forget previous knowledge. 
One of the early incremental approaches was based on the Nearest Class Mean (NCM, \cite{ncm})
classifier, which computes the mean of the feature vectors for
the training samples and each test sample is assigned to the nearest class prototype.
In the more recent literature, there are two main types of approaches. Some methods keep a small memory buffer of previous data in order to replay it while learning new classes \cite{rebuffi-cvpr2017, prabhu2020greedy, ahn2021ss}. Other more complex approaches do not require memory, but exploit extra distillation objectives, constraints on network weights, or rely on generative solutions to reproduce data of previously seen classes \cite{li2017learning,kirkpatrick2017overcoming}. 
Robotics applications for methods of the first family were presented in \cite{icl_robot1,icl_robot2, icl_robot3}. The growing interest in this field is also testified by the ever wider range of datasets designed for it \cite{COSDA_WACV_2021, CORe50, FSIOL}.

Dealing with unconstrained learning conditions means also not knowing a priori which classes will be encountered at test time. \emph{Open Set Recognition} approaches are able to distinguish such unknown objects from the known ones and have been extensively studied both by the computer vision and robotics communities \cite{bendale-boult-cvpr2016,gopenmax,os_robot}

Finally, \emph{Open World Recognition} combines the two previous settings. OWR was first introduced in \cite{bendale2015towards} which proposed NNO, a simple extension of the standard NCM strategy including an unknown rejection policy. Despite its significance for real applications, few other works followed the mentioned seminal paper: 
DeepNNO \cite{mancini2019knowledge}, the deep version of NNO, and B-DOC \cite{fontanel2020boosting} that includes clustering objectives and class-specific rejection thresholds.

\subsection{Domain-Shift}
The discrepancy between training and test data is one of the most relevant problems in robotics where many processes are designed completely in simulated environments, but then need to be applied in the real world \cite{sim2real,selfsup-sim2real,LOAD_ICRA,RAL-vrgoggles}. 
Neglecting the domain shift between source (labeled training) and target (unlabeled test) data leads to brittle methods and frequent fails.
In \emph{Domain Adaptation} the target is available during training and is used to adapt the source model \cite{csurka2017domain}. Some online variants \cite{d2020one} take advantage of a stream of incoming target samples at training time: in this case, there is less available target information and the adaptation process becomes slower \cite{wulfmeier2018,kitting}. 
In \emph{Domain Generalization} the target domain is not known while training on the source: this scenario requires to build models that are robust to any deployment conditions. 
This goal is obtained
{by an effective exploitation of available source data, often} considering to have access to multiple different source domains. 
Most DA and DG works consider the exact same class set shared by source and target, while more recently open-set DA was studied in \cite{liu2019separate,feng2019attract,Pan_2020_CVPR,bucci_hymos_wacv}. Only one work focused on open-set DG under the assumption of multiple available source domains \cite{shu2021open}. 
Overall the main existing adaptive approaches can be organized in few groups: discrepancy-based and feature alignment methods \cite{long2015learning,motiian2017unified}, adversarial learning \cite{ganin2016domain}, data augmentation techniques \cite{volpi2019addressing,nam2021reducing,borlino2021rethinking}, meta-learning \cite{Li2018MLDG}, and self-supervised learning based approaches \cite{loghmani2020unsupervised,jigsaw_pami}. 

\subsection{Contrastive Learning}
\label{sec:contrastiverel}
Self-supervised Contrastive Learning aims at maximizing the agreement among multiple augmentations of the same sample while pushing different instances far apart. This strategy has been recently implemented with different variants \cite{chen2020simple,he2020momentum,Wang_2021_CVPR}: despite not relying on any annotation, the learned hyperspherical embedding (L2 normalized features) captures reliable semantic information, as demonstrated by the effectiveness of the contrastive models used as pretraining for various downstream tasks \cite{jaiswal2021survey}.
The merits of contrastive learning become even more evident when considering its supervised formulation \cite{khosla2020supervised}, simply obtained by exploiting class labels to build positive and negative sample pairs. Samples of the same class are encouraged to cluster together regardless of their intra-class appearance variation, with also large inter-class distances. As a consequence, samples belonging to novel classes end up in low-density regions and measuring the distance to the closer class prototype provides useful rejection information.
This approach has been used in several tasks going from novelty detection \cite{tack2020csi} to cross-domain generalization \cite{zhang2021unleashing},
{open-set domain adaptation} 
\cite{bucci_hymos_wacv}, and class incremental learning \cite{Mai_2021_CVPR}. 

\vspace{2mm}
Previous work has only scratched the surface of the challenging problem of Cross-Domain Open World Recognition:  existing DA and DG approaches have been applied to open world settings to evaluate their potentialities and limits \cite{COSDA_WACV_2021,fontanel_owr_cross_domain}. In this work, we design instead a tailored method for CD-OWR by leveraging supervised contrastive learning.

\section{METHOD}
\subsection{Notations and problem setting}
Let us start from the initial training set $S_0 = \{ \mathbf{x}_i,y_i \} _{i=0}^{N_{S_0}} $ where $\mathbf{x}_i \in \mathbb{X}$ is the image and $y_i$ the corresponding category label from the label set $Y_{S_0}$. We assume to subsequently receive new incremental \textit{tasks} $\{S_1,\ldots,S_K\}$ such that their label sets don't overlap \ie $Y_{S_{k}} \cap Y_{S_{k'}} = \emptyset $  $ \forall \; k,k' \in \left [  0,\ldots,K \right ] $ and $k\neq k'$.  
We call \textit{source domain} the entire labeled training set $ S = \{ \bigcup_{k=0}^{K} S_k \}$ that is drawn from data distribution $p_{\textit{sr}}$. The unlabeled test set  $T = \{ \mathbf{x}_i \}_{i=0}^{N_T} $ is not seen during training and is drawn from the target distribution $p_{\textit{tg}}$, with $p_{\textit{sr}}\neq p_{\textit{tg}}$. Moreover, the target contains both known and unknown categories.
After training on each source task, the goal is to predict for a target test sample whether it is from one of the learned categories or it is unknown. More precisely, our goal is to train a function $f:\mathbb{X} \rightarrow \{Y_s \cup u\}$ such that the target sample is mapped either to one of the semantic categories learned until the current task $Y_s = \{ \bigcup_{k=0}^{t} Y_{S_k} \} $ or to the unknown class $u$. 
We consider $f$ made by three components: a feature extractor  $g: \mathbb{X} \rightarrow \mathbb{Z} $ that maps each image in the feature space, a scoring function $\eta: \mathbb{Z} \rightarrow \mathbb{R}^{|Y_s|} $ that maps the features to a score vector representing the probability that the sample is from one of the learned categories, and finally $ \omega: \mathbb{R}^{|Y_s|} \rightarrow \{Y_s \cup u\} $ that will make the final prediction.

\subsection{Supervised Contrastive Learning} 
We propose to train a contrastive model to obtain a highly structured feature space $\mathbb{Z}$ ready to accommodate the new categories that sequentially arrive with each task. 
The self-supervised contrastive learning loss \cite{chen2020simple,he2020momentum} models the feature space by maximizing the similarity between each instance and its augmented version (\textit{positive pair}), while minimizing the similarity between two different instances (\textit{negative pair}).
In particular, for each training sample $\{ \mathbf{x}_j,y_j \}$, an augmented version $\{ \mathbf{x}_j',y_j \}$ is produced through standard transformations
(\eg grayscale, random crop, color jittering), 
doubling the original batch  $B = \{j=1, \ldots , 2J \}$. 
In our setting we have the category labels of training data, hence we use a \emph{supervised} version of the contrastive learning approach \cite{khosla2020supervised}, which simply modifies the training objective by exploiting sample labels to create positive (same class) and negative (different classes) pairs. 

We consider the feature extractor $g$ composed of an Encoder $E$ and a Projection head $P$. For each sample in $B$, we obtain the representation $\mathbf{z}_j = g(\mathbf{x}_j) = E(P(\mathbf{x}_j))$. 
The final learning objective that we use for the training is:
\begin{equation}
    \sL_{SupClr} = \sum_{j=1}^{2J} \frac{-1}{|\pi(j)|} \sum_{j'\in \pi(j)} \log \frac{\exp(\sigma(\mathbf{z}_j, \mathbf{z}_{j'})/\tau_e)}{\sum\limits_{n\in \nu(j)} \exp(\sigma(\mathbf{z}_j,\mathbf{z}_{n})/\tau_e)}~.
    \label{eq:supclr}
\end{equation}
Here $\nu(j)=B\setminus \{j\}$ is the double batch without the \emph{anchor} sample of index $j$, and $\pi(j) = \{j' \in \nu(j): y_{j'}=y_j\}$ is the set of all the positive pairs. Finally $\tau_e\in \reals^+$ is a temperature parameter, and $\sigma(\cdot, \cdot)$ is the cosine similarity. Thanks to the L2-normalization of the cosine similarity we learn features that lay on a hyperspherical surface and we keep this embedding geometry for all the steps of our learning procedure.

\subsection{Feature space structure and statistics}
The objective described above pushes the data to form compact and well-separated class clusters on the surface of the hypersphere \cite{pmlr-v119-wang20k}.
This structure allows to easily compute some statistics about data distribution, as done in \cite{bucci_hymos_wacv}. First of all we define the prototype of each known class $y_s \in Y_s$ by computing the corresponding feature average ${\textbf{h}_{y_s}=\frac{1}{|y_s|}\sum_{k\in y_s}\textbf{z}_{k}}\,,$ re-projected on the unit hypersphere. 
Here $|y_s|$ indicates the number of samples of class $y_s$ and $k$ is the index that runs on all them.
We also define the angular distance measure $d_a(\textbf{z}_i, \textbf{z}_j)=\{1-\sigma_{[0,1]}(\textbf{z}_i,\textbf{z}_j)\}$ which rescale the cosine similarity in $[0,1]$ and translates it.
We are interested in two feature space statistics:
\begin{itemize}
    \item the \emph{class sparsity}, which measures the average distance between a prototype $\textbf{h}_{y_s}$ and the nearest one among the others $\textbf{h}_*$: $\theta = \frac{1}{|Y_s|}\sum_{y_s\in Y_s} d_a(\textbf{h}_*,\textbf{h}_{y_s})$;
    \item the \emph{class compactness}, which measures the average distance between training samples and the corresponding class prototypes: $\phi = \frac{1}{|Y_s|}\sum_{y_s\in Y_s}\left\{ \frac{1}{|y_s|} \sum_{k\in y_s} d_a(\textbf{h}_{y_s},\textbf{z}_k)\right\}$
\end{itemize}
\noindent These two metrics describe the data distribution and can be used as reference to make decisions both on the stopping criterion for each learning episode, and on the known-unknown class separation. 

\subsection{Incremental protocol}
To keep the focus on the effectiveness of the supervised contrastive loss function for the CD-OWR task, we do not implement any complex 
additional module to avoid forgetting in the incremental procedure. 
We, adopt only two simple strategies: i) as done by other incremental learning algorithms we keep a (limited- and fixed-size) replay buffer containing a subselection of samples from previous tasks; ii) we perform class balancing at the batch level, by putting in each training mini-batch at least two samples of each class. 
We expect our learning objective to manage the data and progressively make room on the hyperspherical feature space to accommodate new classes while exploiting replay samples to maintain 
reserved space for the old ones.

\subsection{Stop-training criterion}
Intuitively class clusters in our feature space cannot be well separated if $\theta < 2 \phi$. In fact, we can see $\phi$ as a measure of the radius of clusters: if the distance between two class centroids is lower than the sum of their radii the two clusters will inevitably overlap. In this case, samples of the two classes cannot be distinguished. Moreover, with no \emph{empty space} between class clusters, there's no space for unknown data either. In order to avoid this condition, we can impose a constraint on the quality of the feature space for our output model. Given that the compactness and separation of the clusters increase during training, we can enforce the described relation between $\theta$ and $\phi$ by using a specific stopping criterion in the learning procedure. We consider each learning task as converged only when
\begin{equation}
    \lambda > 1 + \varepsilon, \quad \textit{with} \quad \lambda = \cfrac{\theta}{2 \phi} \quad \textit{and} \quad \varepsilon \geq 0~.
    \label{eq:stop_criterion}
\end{equation}
Here $\varepsilon$ can be seen as a \emph{minimum desired margin} between two clusters. 

\subsection{Threshold definition}
By following the same logic of the NCM strategy, our scoring function is based on the distance of each target sample to its nearest source class prototype: $\eta(\textbf{z}^t) =
d_a(\textbf{h}_{{y}_s},\textbf{z}^t)$.
This approach is particularly suitable in the OWR setting as it enables the use of a threshold on the distance to perform known-unknown separation. Therefore we can define our prediction function $\omega$ for target samples as:
\begin{equation}
    \hat{y}^t = \omega(\textbf{z}^t) = 
    \begin{cases}
        \argmin_{y_s}(d_a(\textbf{h}_{y_s},\textbf{z}^t)) & \mbox{if } \min{_{y_s}(d_a(\textbf{h}_{y_s},\textbf{z}^t))}< \tau\\
        \texttt{unknown} & \mbox{otherwise} 
    \end{cases}
\end{equation}

\begin{figure}[t!]
\vspace{2mm}
\centering
\includegraphics[width=0.48\textwidth]{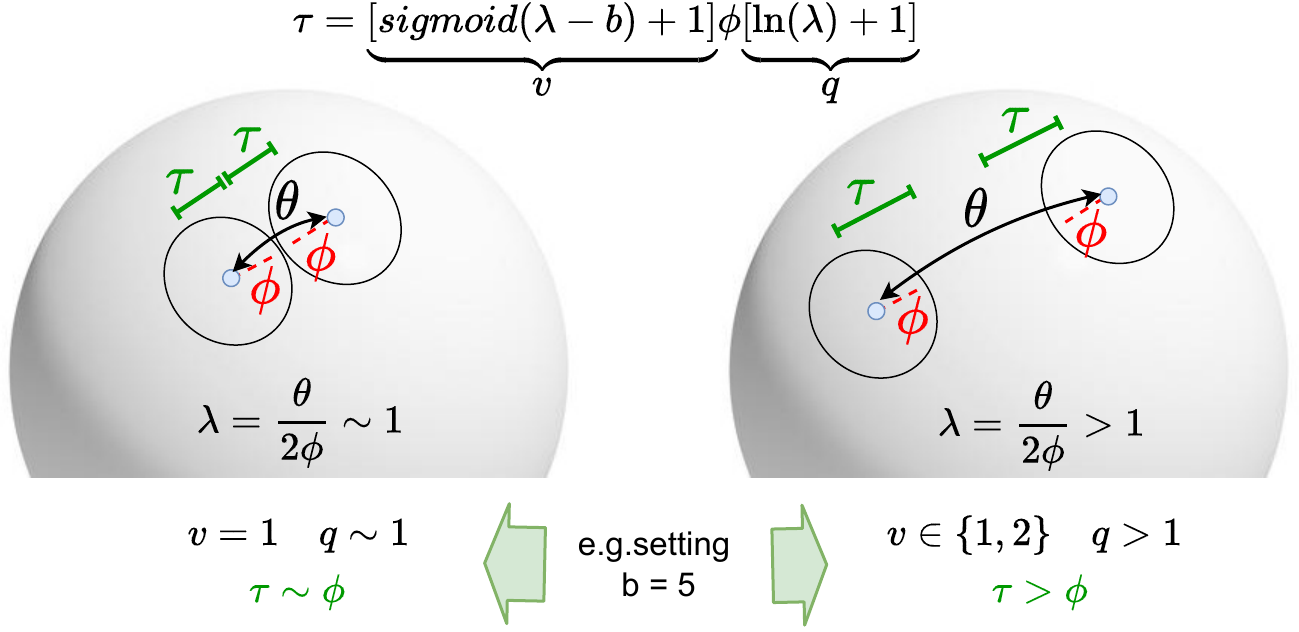}\vspace{-2mm}
    \caption{Visualization of two data clusters on the learned hyperspherical embedding and of the corresponding threshold value. Tuning $b$ means tuning $v$ in $\{1,2\}$. See sec. \ref{sec:ablation} for a detailed discussion.}\vspace{-5mm}
    \label{fig:sphere}
\end{figure}

The value of the threshold $\tau$ is one of the most important choices for an open world approach  (see Table I in \cite{fontanel_owr_cross_domain}). In our case, we can take advantage of the feature space statistics to define the threshold.
Inspired by \cite{bucci_hymos_wacv}, we set
\begin{equation}
    \tau =   [\textit{sigmoid} \left( \lambda - b\right) + 1]  \cdot \phi \cdot \left( \ln\left(\lambda\right) +1 \right)
    \label{eq:threshold}
\end{equation}
Differently from \cite{bucci_hymos_wacv} we have the constraint imposed through Eq. (\ref{eq:stop_criterion}), which makes the result of the logarithm always positive. We also propose a parametric formulation for the first term (between squared brackets) thanks to which we obtain a good known-unknown balancing in all the tested benchmarks (see Fig. \ref{fig:sphere}).

\section{Experimental Setup}
\subsection{Datasets and experimental protocol}
To assess the performance of \OUR, we rely on the benchmark proposed in \cite{fontanel_owr_cross_domain}, by also extending it in order to consider more recent literature and new data collections. In particular we focus on five datasets perfectly suited for our scope. All of them contain daily-life objects (spanning from fruit and vegetables to tools and containers) recorded under very different acquisition conditions.
\par
\textbf{RGB-D Object dataset} (ROD) \cite{lai2011large} is one of the most used RGB-D dataset in robotics for object categorization. The objects were placed on a table and captured with different viewpoints. The recording was done in a strictly controlled environment without any source of noise \ie without clutter, with a fixed illumination and background. 
\par
\textbf{Synthetic ROD} (synROD) \cite{loghmani2020unsupervised} is the synthetic version of ROD, proposed to analyze
the synthetic-to-real domain shift problem in a robotic context. It was recorded using publicly available 3D models rendered using a ray-tracing engine in Blender to simulate photorealistic lighting.   
\par
\textbf{Autonomous Robot Indoor Dataset} (ARID) \cite{loghmani2018recognizing} is a challenging dataset in which the objects were captured in a cluttered environment: the same object appears with several backgrounds, scales, views, lighting conditions, and different levels of occlusions. The purpose of this dataset was to evaluate the robustness of a recognition model when dealing with difficult but realistic scenarios.
\par
\textbf{Continual Open Set Domain Adaptation for Home Robot} (COSDA-HR) \cite{COSDA_WACV_2021} 
is a dataset composed of a source domain with hand-held objects placed in front of a uniform background and a target domain with objects captured in various natural locations in a home environment.
\par 
\textbf{Continuos Object Recognition 50} (CORe50)
is a collection of photos of domestic objects, captured while being held by the operator in 11 distinct sessions (8 indoor and 3 outdoor).
\par
For what concern ROD, synROD and ARID our experimental protocol follows the same configuration proposed in \cite{fontanel_owr_cross_domain}: among the 51 object categories that they share, we randomly consider 26 of them as known and 25 as unknown; we start with 11 known categories that increase by 5 at each incremental step for a total of four sequential tasks. 
For COSDA-HR instead, we follow \cite{COSDA_WACV_2021}: the dataset is composed of 160 categories incrementally learned 10 at a time for a total of 16 tasks. The dataset includes a single unknown category made by a heterogeneous set of objects. 
CORe50 was designed to perform instance classification on 50 objects. We consider 10 of them in the first learning episode and add 5 in each of the subsequent three, keeping the last 25 as unknown. We consider the indoor (train data) $\to$ outdoor (test data) domain shift.
To better assess the performances of the methods, for all the experiments, we consider five different random class orders and we report the obtained average. 
\par
\textbf{Metrics.} For the evaluation we use the same metrics used in \cite{fontanel_owr_cross_domain}.  
\textit{Acc} (Accuracy) measures the ability of the model to correctly predict the categories of the known target samples. 
\textit{Acc-WR} (Accuracy Without Rejection) is similar to Acc, but the accuracy is computed without rejecting the target samples identified as unknown. 
\textit{OWR-H} (Open World Harmonic Mean) evaluates the performance of the model as a whole, it is the harmonic mean between Acc-WR and the model's accuracy in unknown sample detection.

\subsection{Competitors}
We follow \cite{fontanel_owr_cross_domain} comparing \OUR against state-of-the-art methods in OWR, enhanced with single-source DG approaches to deal with the domain-shift. As competitors for the OWR setting we consider: \textbf{NNO} \cite{bendale2015towards} a non-parametric approach that exploits the Nearest-Class Mean (NCM) algorithm \cite{mensink2012metric} to compute the class centroids with the features extracted from a pretrained deep architecture; its more advanced version \textbf{DeepNNO} \cite{mancini2019knowledge}, in which the feature extractor is end-to-end trained and the rejection threshold is not fixed but updated during training, and \textbf{B-DOC} \cite{fontanel2020boosting} which includes two clustering constraints in the optimization process and proposes a class-specific rejection threshold. 
We also include the state-of-the-art CIL method 
\textbf{SS-IL} \cite{ahn2021ss} that uses separate softmax output layers combined with
task-wise knowledge distillation to mitigate the bias toward the new classes. 
For both OWR and CIL the training happens in sequence on multiple tasks.
The main difference is that, in CIL, the evaluation step is done on a test set composed only of images from categories learned up to the current task, in OWR the test set contains also samples from categories not learned (yet): at test time the model has to reject the unknown samples while assigning a class label only to those belonging to the learned categories. 
As a consequence, in CIL there is no need for 
the known/unknown separation threshold. 
To adapt SS-IL to the  OWR scenario we exploit the Maximum Softmax Probability \cite{hendrycks2016baseline} of the predictions vector and to choose the threshold we rely on the logic proposed in \cite{lakshminarayanan2017simple}.

We follow \cite{fontanel_owr_cross_domain} also for the DG literature: a data augmentation based technique \textbf{RSDA} \cite{volpi2019addressing}, a self-supervised based technique \textbf{RR} \cite{loghmani2020unsupervised} and a method based on a regularization strategy \textbf{SC} \cite{huang2020self}. 
Moreover, we include a very recent DG approach \textbf{SagNet} \cite{nam2021reducing} that disentangles the sample content and style to let the network focus more on the first than on the second.

\begin{table*}[t]
\vspace{2mm}
\caption{Results (\%) averaged over five random class orders.}
\resizebox{\textwidth}{!}{

\begin{tabular}{|@{~}c@{}c@{~}|}
\hline
\multirow{2}{*}{\textbf{OWR/CIL}}  & \multirow{2}{*}{\hspace{-10pt}\textbf{DG}} \\
& \\
\hline

{NNO  \cite{bendale2015towards} } & {\multirow{4}{*}{-}} \\
{DeepNNO} \cite{mancini2019knowledge} &  \\
{B-DOC \cite{fontanel2020boosting}} & \\
{SS-IL \cite{ahn2021ss} }  &  \\
      
\hline
\hline

{NNO  \cite{bendale2015towards} } &  {\multirow{4}{*}{\texttt{+} RR \cite{loghmani2020unsupervised}}} \\
{DeepNNO} \cite{mancini2019knowledge} & \\
{B-DOC \cite{fontanel2020boosting}} & \\
{SS-IL \cite{ahn2021ss} }  & \\
      
\hline

{NNO  \cite{bendale2015towards} } & {\multirow{4}{*}{\texttt{+} SC \cite{huang2020self}}} \\
{DeepNNO} \cite{mancini2019knowledge} & \\
{B-DOC \cite{fontanel2020boosting}} & \\
{SS-IL \cite{ahn2021ss} }  &\\
      
\hline
{NNO  \cite{bendale2015towards} } & {\multirow{4}{*}{\texttt{+} RSDA \cite{volpi2019addressing}}} \\
{DeepNNO} \cite{mancini2019knowledge} & \\
{B-DOC \cite{fontanel2020boosting}} & \\
{SS-IL \cite{ahn2021ss} }  & \\

\hline
{NNO  \cite{bendale2015towards} } & {\multirow{4}{*}{\texttt{+} SagNet \cite{nam2021reducing}}} \\
{DeepNNO} \cite{mancini2019knowledge} &\\
{B-DOC \cite{fontanel2020boosting}} &\\
{SS-IL \cite{ahn2021ss} }  & \\
\hline
\hline
\multicolumn{2}{|c|}{\textbf{\OUR}}   \\
\hline
\end{tabular}

\begin{tabular}{|@{~}c@{~}|@{~}c@{~}|@{~}c@{~}|}

\hline
 \multicolumn{3}{|@{~}c@{~}|}{\textbf{ROD $\rightarrow$ ARID }} \\
\hline

{Acc-WR} & { Acc } & {OWR-H}\\

\hline

{18.4}  & {3.1} & 5.9 \\
{21.3} &	{7.3}  & 13.4 \\
 {22.3}  & {10.0} & 17.5\\  
  {17.6}  & {14.7}  & {16.4}  \\ 
 \hline
 \hline
 
{27.1}  & {13.6} & 21.7 \\
	{33.5} &	{16.0}  & 25.8 \\
 {32.2}  & {11.7} & 20.4\\  
   {17.3}  & {13.0}  & {17.9}  \\ 
 \hline
 
	{14.1}  &{9.8} & 15.5 \\
	{20.9} &	{15.9}  & 22.0 \\
 {19.6}  & {13.1} & 20.4 \\  
   {15.2}  & {12.9}  & {14.7}  \\ 
 \hline
 
{25.0}  & {12.8} & 20.7 \\
	{33.3} &	{14.9}  & 24.6 \\
{31.9}  & {12.2} & 21.1\\  
   {29.9}  & {\textbf{24.4}}  & {24.1}  \\ 
 \hline
 
	{19.1}  & {3.9} & 7.4 \\
	{22.5} &	{8.7}  & 15.5 \\
 {23.7}  & {10.7} & 18.2\\  
   {24.9}  & {19.4}  & {21.8}  \\ 
 \hline
 \hline
     {\textbf{34.0}}  & {18.8}  & {\textbf{28.6}} \\
 \hline
\end{tabular}

\begin{tabular}{|@{~}c@{~}|@{~}c@{~}|@{~}c@{~}|}
\hline
\multicolumn{3}{|@{~}c@{~}|}{\textbf{synROD $\rightarrow$ ARID }}        \\
\hline
{Acc-WR} & { Acc } & {OWR-H}\\
\hline
{16.2}  & {7.8}  & {13.7} \\
{15.9} & {5.4} &	{10.0}  \\
{16.5}  &	{2.2} &	{4.3} \\
{21.3}  &	{9.6}   &	{16.1}   \\ 
\hline
\hline
{15.8}  & {7.2}  & {12.5} \\
{14.2} & {4.9} &	{9.3}  \\
{15.7}  &	{2.2} &	{4.3} \\
{19.7}  &	{6.6}   &	{11.6}   \\ 
\hline
{16.0}  & {11.6}  & {16.9} \\
{15.5} & {8.4} &	{14.6}  \\
{16.5}  &	{10.0} &	{16.1} \\
{19.0}  &	{7.9}   &	{13.2}   \\ 
\hline

{16.3}  & {8.6}  & {14.4} \\
{15.3} & {4.2} &	{8.0}  \\
{16.3}  &	{2.5} &	{4.9} \\
{20.3}  &	{7.6}   &	{12.8}   \\ 
\hline

{15.2}  & {7.3}  & {12.7} \\
{13.7} & {4.7} &	{8.8}  \\
{18.2}  &	{4.6} &	{8.5} \\
{20.9}  &	{8.9}   &	{14.9}   \\ 
\hline
\hline
{\textbf{29.8}}  &	{\textbf{21.3}}   &	{\textbf{28.1}} \\

\hline
\end{tabular}

\begin{tabular}{|@{~}c@{~}|@{~}c@{~}|@{~}c@{~}|}

\hline
\multicolumn{3}{|@{~}c@{~}|}{\textbf{synROD $\rightarrow$ ROD}}        \\
\hline
{Acc-WR} & { Acc } & {OWR-H}\\
\hline
{21.3}  &	{13.3}  &	{21.1}  \\
{24.4}   &	{9.6}   &	{17.0} \\
{27.6} &{5.2}  &		{9.9} \\ 
{29.3} &	  	{16.9}  &	{22.9} \\
\hline
\hline
{25.9}  & {17.1}  & {23.8} \\
{34.1} & {15.4} &	{25.2}  \\
{35.9}  &	{9.7} &	{17.3} \\
{30.9}  &	{16.2}   &	{23.7}   \\ 
\hline
{21.9}  & {18.8}  & {21.2} \\
{25.9} & {17.0} &	{25.3}  \\
{26.7}  &	{18.0} &	{23.2} \\
{26.8}  &	{14.6}   &	{21.0}   \\ 
\hline

{26.7}  & {18.4}  & {24.5} \\
{34.2} & {14.0} &	{23.5}  \\
{37.9}  &	{10.8} &	{19.1} \\
{\textbf{38.7}}  &	{23.6}   &	{30.3}   \\ 
\hline

{20.3}  & {12.4}  & {19.4} \\
{17.9} & {7.1} &	{12.8}  \\
{28.9}  &	{9.1} &	{16.1} \\
{32.8}  &	{17.7}   &	{24.5}   \\ 
\hline
\hline
{34.1}  &	{\textbf{24.0}}   &	{\textbf{30.7}} \\

\hline

\end{tabular}

\begin{tabular}{|@{~}c@{~}|@{~}c@{~}|@{~}c@{~}|}

\hline
\multicolumn{3}{|@{~}c@{~}|}{\textbf{COSDA-HR}}        \\
\hline
{Acc-WR} & { Acc } & {OWR-H}\\
\hline
{8.2}  &	{3.7}  &	{7.0}  \\
{15.1}   &	{8.2}   &	{12.8} \\
{13.2} &{0.7}  &		{1.3} \\ 
{8.4} &	  	{5.0}  &	{6.2} \\
\hline
\hline
{8.2}  & {3.2}  & {6.5} \\
{13.8} & {6.9} &	{11.0}  \\
{12.3}  &	{0.5} &	{1.0} \\
{7.8}  &	{1.4}   &	{2.6}   \\ 
\hline
{6.4}  & {4.6}  & {7.2} \\
{15.3} & {11.7} &	{15.5}  \\
{13.0}  &	{1.9} &	{3.3} \\
{9.0}  &	{5.5}   &	{6.6}   \\ 
\hline

{8.9}  & {2.1}  & {3.9} \\
{18.4} & {10.9} &	{17.1}  \\
{18.2}  &	{0.6} &	{1.0} \\
{17.8}  &	{8.3}   &	{12.6}   \\ 
\hline

{8.1}  & {3.2}  & {6.4} \\
{8.5} & {3.9} &	{7.2}  \\
{11.3}  &	{0.5} &	{1.1} \\
{8.3}  &	{3.7}   &	{6.1}   \\ 
\hline
\hline
{\textbf{20.1}}  &	{\textbf{16.2}}   &	{\textbf{21.4}} \\

\hline

\end{tabular}

\begin{tabular}{|@{~}c@{~}|@{~}c@{~}|@{~}c@{~}|}

\hline
\multicolumn{3}{|@{~}c@{~}|}{\textbf{\textcolor{black}{CORe50}}}        \\
\hline
{Acc-WR} & {Acc} & {OWR-H}\\
\hline
{15.0}  &	{0.6}  &	{1.2}  \\
{17.0}   &	{4.5}   &	{8.1} \\
{15.7} & {2.2}  &	{3.8} \\ 
{23.2} & {16.8}  &	{18.6} \\
\hline
\hline
{14.5}  & {0.4}  & {0.9} \\
{17.5} & {5.0} &	{8.9}  \\
{19.5}  &	{4.5} &	{7.1} \\
{20.2}  &	{9.7}   &	{12.8}   \\ 
\hline
{13.3}  & {2.9}  & {5.3} \\
{18.2} & {8.8} &	{13.6}  \\
{17.1}  &	{4.7} &	{6.8} \\
{19.3}  &	{14.7}   &	{16.0}   \\ 
\hline

{22.1}  & {13.1}  & {16.1} \\
{38.0} & {20.8} &	{30.7}  \\
{\textbf{41.4}}  &	{9.8} &	{15.9} \\
{38.1}  &	{\textbf{25.9}}   &	{30.6}   \\ 
\hline

{15.9}  & {2.5}  & {4.7} \\
{19.3} & {7.6} &	{12.4}  \\
{17.3}  &	{3.6} &	{5.9} \\
{27.3}  &	{17.6}   &	{23.5}   \\ 
\hline
\hline
{33.9}  &	{23.9}   &	{\textbf{32.9}} \\

\hline

\end{tabular}

}
\vspace{-3mm}
\label{tab:sota}
\end{table*}

\subsection{Implementation Details}
We implement \OUR considering the same protocol adopted for all our competitors: we use a ResNet18 backbone trained from scratch using images of size $64\times64$. When learning a new task we keep a fixed-size memory buffer to store $M = 2000$ samples of the classes of previous tasks by choosing them randomly.  We train each task until our stop training condition (Eq. \ref{eq:stop_criterion}) is matched. 
\OUR has only two hyperparameters, $\varepsilon$ and $b$, and is robust to their value as discussed in Sec. \ref{sec:ablation}.

\section{Experiments}

In this section, we report the results obtained evaluating \OUR against the competitors introduced before. 
We consider different levels of domain-shift 
between source and target domain (we use the notation \textit{source} $\to$ \textit{target}).
We show how existing OWR and CIL solutions are far from solving the task of Cross-Domain Open World Recognition even if enhanced with DG approaches to bridge the domain gap. Instead, COW with a single loss and without any additional module to mitigate the domain shift outperforms the current state-of-the-art. 

\definecolor{color_our}{rgb}{0.39, 0.58, 0.93}
\definecolor{color_nno}{rgb}{1.0, 0.0, 0.0}
\definecolor{color_dnno}{rgb}{1.0, 0.85, 0.0}
\definecolor{color_bdoc}{rgb}{0.0, 0.85, 0.0}
\definecolor{color_ssil}{rgb}{1.0, 0.5, 0.0}

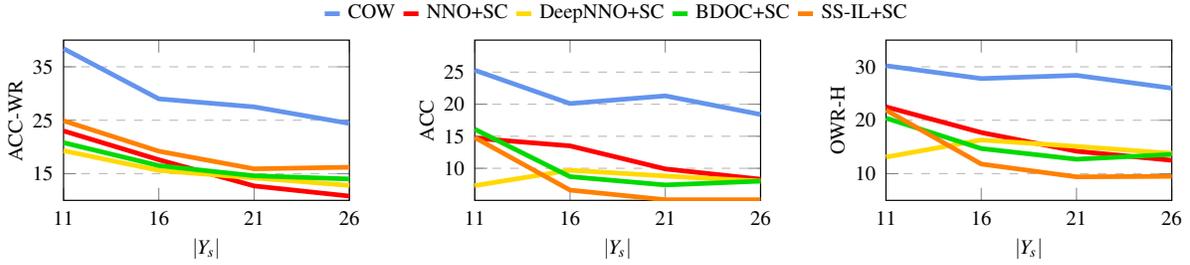
\begin{figure*}[h]
\centering
\begin{minipage}[t]{0.9\textwidth}

    \resizebox {\columnwidth} {!} {
        \begin{tikzpicture}
        
        \tikzstyle{every node}=[font=\scriptsize]
        \pgfplotsset{scaled x ticks=false,every axis legend/.append style={
at={(0.5,1.03)},
anchor=south}}

        \begin{axis}[
          name=left,
          enlargelimits=false,
          x label style={at={(axis description cs:0.5,0.1)},anchor=north},
          y label style={at={(axis description cs:0.13,0.5)},anchor=north},
          ylabel={ACC-WR},
          xlabel={$|Y_s|$},
           xmin=11, xmax=26,
           ymin=10, ymax=40,
          xtick={11,16,21,26},
          ytick={15,25,35},
          ymajorgrids=true,
          grid style=dashed,
          width=5cm,
          height=3.5cm,
          legend columns=-1,
        every axis plot/.append style={ultra thick},
        every mark/.append style={mark size=50pt},
        label style={font=\scriptsize},
        legend style={font=\scriptsize},
        legend image post style={scale=0.3}
        ]

        \addplot[color=color_our,mark=none]
        coordinates {(11, 38.4)(16, 29.0)(21, 27.5)(26, 24.4)};
        \addlegendentry{\OUR}
        
        \addplot[color=color_nno,mark=none]
        coordinates {(11, 23.0)(16, 17.6)(21, 12.7)(26, 10.8)};
        \addlegendentry{NNO+SC}
        
        \addplot[color=color_dnno,mark=none]
        coordinates {(11, 19.3)(16, 15.6)(21, 14.2)(26, 12.8)};
        \addlegendentry{DeepNNO+SC}
        
        \addplot[color=color_bdoc,mark=none]
        coordinates {(11, 20.8)(16, 16.5)(21, 14.6)(26,14.0)};
        \addlegendentry{BDOC+SC}
        
        \addplot[color=color_ssil,mark=none]
        coordinates {(11, 24.9)(16, 19.2)(21, 15.9)(26, 16.2)};
        \addlegendentry{SS-IL+SC}
        
        \legend{}

 \end{axis}
        \begin{axis}[
          name=center,
          at={(left.south east)},
          xshift=1.5cm,
          enlargelimits=false,
          x label style={at={(axis description cs:0.5,0.1)},anchor=north},
          y label style={at={(axis description cs:0.13,0.5)},anchor=north},
          ylabel={ACC},
          xlabel={$|Y_s|$},
          xmin=11, xmax=26,
          ymin=5, ymax=30,
          xtick={11,16,21,26},
          ytick={10,15,20,25},
          ymajorgrids=true,
          grid style=dashed,
          width=5cm,
          height=3.5cm,
          legend columns=-1,
        every axis plot/.append style={ultra thick},
        every mark/.append style={mark size=50pt},
        label style={font=\scriptsize},
        legend style={font=\scriptsize, draw=none},
        legend image post style={scale=0.3}
        ]

        \addplot[color=color_our,mark=none]
        coordinates {(11, 25.3)(16, 20.1)(21, 21.3)(26, 18.4)};
        \addlegendentry{\OUR}
        
        \addplot[color=color_nno,mark=none]
        coordinates {(11, 14.7)(16, 13.5)(21, 9.9)(26, 8.3)};
        \addlegendentry{NNO+SC}
        
        \addplot[color=color_dnno,mark=none]
        coordinates {(11, 7.3)(16, 9.7)(21, 8.8)(26, 8.0)};
        \addlegendentry{DeepNNO+SC}
        
        \addplot[color=color_bdoc,mark=none]
        coordinates {(11, 16.1)(16, 8.7)(21, 7.4)(26, 8.0)};
        \addlegendentry{BDOC+SC}
        
        \addplot[color=color_ssil,mark=none]
        coordinates {(11, 14.8)(16, 6.6)(21, 5.1)(26, 5.1)};
        \addlegendentry{SS-IL+SC}

        \end{axis}
        
        \begin{axis}[
          at={(center.south east)},
          xshift=1.5cm,
          enlargelimits=false,
          x label style={at={(axis description cs:0.5,0.1)},anchor=north},
          y label style={at={(axis description cs:0.13,0.5)},anchor=north},
          ylabel={OWR-H},
          xlabel={$|Y_s|$},
          xmin=11, xmax=26,
          ymin=5, ymax=35,
          xtick={11,16,21,26},
          ytick={10,20,30},
          ymajorgrids=true,
          grid style=dashed,
          width=5cm,
          height=3.5cm,
          legend columns=-1,
        every axis plot/.append style={ultra thick},
        every mark/.append style={mark size=50pt},
        label style={font=\scriptsize},
        legend style={font=\scriptsize},
        legend image post style={scale=0.3}
        ]

        \addplot[color=color_our,mark=none]
        coordinates {(11, 30.2)(16, 27.8)(21, 28.4)(26, 26.0)};
        
        \addplot[color=color_nno,mark=none]
        coordinates {(11, 22.5)(16, 17.7)(21, 14.2)(26, 12.5)};
        
        \addplot[color=color_dnno,mark=none]
        coordinates {(11, 13.1)(16, 16.3)(21, 15.1)(26, 13.8)};
        
        \addplot[color=color_bdoc,mark=none]
        coordinates {(11, 20.4)(16, 14.7)(21, 12.7)(26, 13.6)};
        
        \addplot[color=color_ssil,mark=none]
        coordinates {(11, 21.9)(16, 11.8)(21, 9.4)(26, 9.5)};

        \end{axis}
        
        \end{tikzpicture}%
    }\vspace{-3mm}
\caption{Performance analysis at subsequent learning episodes for synROD $\to$ ARID. The number of known classes $|Y_s|$ increases and the plots show how \OUR maintains a consistent gain over all the competitors.}
\label{fig:incremental_trend}
\vspace{-2mm}
\end{minipage}
\end{figure*}

\subsection{Results}
In the upper part of Table \ref{tab:sota} we report the results obtained using vanilla state-of-the-art OWR and CIL approaches to solve the cross-domain OWR task without the help of DG methods. \textit{Fontanel et al.} \cite{fontanel_owr_cross_domain} already showed how these methods perform poorly in a cross-domain scenario.
The second block of the table presents the results obtained combining the OWR and CIL approaches with single-source DG methods.
We re-ran\footnote{Our results are consistent with those of \cite{fontanel_owr_cross_domain}, made exception for few cases. An issue affected some of the numbers: we identified and corrected the problem via private communications with the authors.} the experiments originally presented in  \cite{fontanel_owr_cross_domain}, also considering the most recent methods SS-IL and SagNet. 
In the last row of the table, we show the results obtained by \OUR, the only approach that handles all the challenges of the CD-OWR setting at once without the need to integrate extra adaptive modules.

We now discuss the results referring to the OWR-H metric that was shown to be the most appropriate one to evaluate an open-set approach \cite{bucci2020effectiveness,fontanel2020boosting}. For what concerns the ROD $\rightarrow$ ARID and synROD $\rightarrow$ ROD domain shifts, as well as the CORe50 dataset, we can generally see an improvement after adding each one of the DG approaches. There are only a few exceptions like SS-IL+SC whose combination may be incompatible: they both exploit the value of the gradient to formulate their solutions, but from two different points of view that might disagree. 
Anyway, the mean improvement gained with the addition of the DG strategies confirms the generalization failure of the existing OWR/CIL approaches that need auxiliary loss functions in order to properly work on a target domain different from the training one. We also observe that among the considered DG strategies the one that most often produces the highest results is the data augmentation-based approach RSDA.
This confirms the great advantage that a strong data augmentation can provide
in knowledge generalization \cite{volpi2019addressing, borlino2021rethinking}. However in some edge cases, it may be not enough: in the synROD $\rightarrow$ ARID shift, and in COSDA-HR dataset, all the considered DG strategies, including RSDA, do not seem to provide a significant and consistent improvement. In these cases, the domain shift is quite severe since it includes a (realistic) target domain whose images have been recorded in a cluttered environment, which is very different from the neat (and possibly synthetic) training set.
Indeed the considered DG strategies are not suited to reduce such a large domain gap.
As Table \ref{tab:sota} clearly shows, \OUR obtains the best results over all the experiments,
proving its effectiveness. 

\subsection{Domain Generalization through Contrastive Learning}
\begin{table}
\caption{Contrastive learning vs Data augmentation.}
\resizebox{1.01\linewidth}{!}{
{
\renewcommand\arraystretch{1.1}
\begin{tabular}{|c@{}c@{}c@{}c@{}c@{}|}

\hline
\multirow{2}{*}{\textbf{OWR/CIL}}  &  \multirow{2}{*}{\hspace{-10pt}\textbf{DG}} &  \multicolumn{3}{|c|}{\textbf{synROD $\rightarrow$ ARID }}        \\
\cline{3-5}
& &  \multicolumn{1}{|c|}{ Acc-WR  } & \multicolumn{1}{c|}{ \hspace{0.1cm} Acc \hspace{0.1cm} }  & \multicolumn{1}{c|}{OWR-H}\\
 \hline
\multicolumn{1}{|c}{NNO \cite{bendale2015towards}}  &  \multicolumn{1}{c}{\texttt{+} SC \cite{huang2020self} + RC}  &	\multicolumn{1}{|c|}{16.5}  &	\multicolumn{1}{c|}{13.8}  &	\multicolumn{1}{c|}{14.6} \\
			
 \hline
 
\multicolumn{1}{|c}{NNO \cite{bendale2015towards}} &  \multicolumn{1}{c}{\multirow{4}{*}{\texttt{+} RSDA \cite{volpi2019addressing} + RC}}  &	\multicolumn{1}{|c|}{17.3}  &	\multicolumn{1}{c|}{12.5}  &	\multicolumn{1}{c|}{14.5} \\

\multicolumn{1}{|c}{DeepNNO \cite{mancini2019knowledge}} &  &	\multicolumn{1}{|c|}{20.4}  &	\multicolumn{1}{c|}{10.7}  &	\multicolumn{1}{c|}{17.7} \\

\multicolumn{1}{|c}{B-DOC \cite{fontanel2020boosting}} & & 	\multicolumn{1}{|c|}{23.8}  &	\multicolumn{1}{c|}{13.3}  &	\multicolumn{1}{c|}{20.1} \\

\multicolumn{1}{|c}{SS-IL \cite{ahn2021ss}} & & 	\multicolumn{1}{|c|}{27.0}  &	\multicolumn{1}{c|}{11.2}  &	\multicolumn{1}{c|}{17.9} \\

\hline
\hline
 
    \multicolumn{2}{|c|}{\textbf{\OUR}}  &	\multicolumn{1}{c|}{\textbf{29.8}}  & \multicolumn{1}{c|}{\textbf{21.3}}  & \multicolumn{1}{c|}{\textbf{28.1}}  \\
\hline
\end{tabular}
}
}
\vspace{-3mm}
\label{tab:ablation}
\end{table}
As described in Sec. \ref{sec:contrastiverel}, contrastive learning relies on data augmentation and the techniques usually adopted to create data variants are the same as RSDA, plus random resized crop (RC).
To understand whether the advantage of COW originates mainly from this augmentation, or from the specific way in which it is used by the contrastive loss, we decided to provide other baselines with the additional RC augmentation procedure.
In Table \ref{tab:ablation} we consider the synROD $\rightarrow$ ARID domain shift. We compare against our best competitor (for this shift) NNO+SC, and the methods that already use a strong augmentation because integrated with RSDA. We can observe that the performance of the considered competitors does not always increase with this addition: clearly, this procedure can also be detrimental to methods that are not designed to manage it. Nevertheless, even when the performance of the competitors increases, \OUR still keeps the best results. This evidences that the data augmentation is not enough and the 
contrastive logic is a fundamental component to enable generalization.

\definecolor{cornflowerblue}{rgb}{0.39, 0.58, 0.93}
\definecolor{my_orange}{rgb}{1.0, 0.5, 0.0}
\begin{figure}[tb]
\centering

\begin{minipage}[t]{0.49\textwidth}
     \resizebox {\columnwidth} {!} {
        \begin{tikzpicture}
        \tikzstyle{every node}=[font=\scriptsize]
        \pgfplotsset{scaled x ticks=false,every axis legend/.append style={
at={(0.5,1.03)},
anchor=south}}

        \begin{axis}[
          name=eps_left,
          enlargelimits=false,
          x label style={at={(axis description cs:0.5,0.1)},anchor=north},
          y label style={at={(axis description cs:0.13,0.5)},anchor=south},
          xlabel={ $ \mathbf{\epsilon}  $},
           xmin=0.1, xmax=0.75,
           ymin=22, ymax=30,
          xtick={0.1,0.25,0.5,0.75},
          ytick={22,24, 26,28, 30},
          ymajorgrids=true,
          grid style=dashed,
          width=5cm,
          height=3.5cm,
          legend columns=-1,
        every axis plot/.append style={ultra thick},
        every mark/.append style={mark size=50pt},
        label style={font=\scriptsize},
        legend style={font=\scriptsize},
        legend image post style={scale=0.3}
        ]
        
        \addplot[color=cornflowerblue,mark=none]
        coordinates {(0.1, 26.4)(0.25, 28.1)(0.5, 28.0)(0.75, 25.9)};
        
        \addplot[color=my_orange,mark=none]
        coordinates {(0.1, 28.6)(0.25, 28.6)(0.5, 28.6)(0.75, 28.6)};

 \end{axis}

        \begin{axis}[
          at={(eps_left.south east)},
          enlargelimits=false,
          xshift=0.5cm,
          x label style={at={(axis description cs:0.5,0.1)},anchor=north},
          y label style={at={(axis description cs:0.13,0.5)},anchor=south},
          xlabel={ $ b  $},
           xmin=3, xmax=17,
           ymin=22, ymax=30,
          xtick={3,5,7,9,11,13,15},
          ytick={22,24,26,28,30},
          ymajorgrids=true,
          grid style=dashed,
          width=5cm,
          height=3.5cm,
          legend columns=-1,
        legend style={at={(-0.1,1.05)},draw=none},
        every axis plot/.append style={ultra thick},
        every mark/.append style={mark size=50pt},
        label style={font=\scriptsize},
        legend style={font=\scriptsize},
        legend image post style={scale=0.3}
        ]
        
        \addplot[color=cornflowerblue,mark=none]
        coordinates {(3, 26.4)(4, 27.9)(5, 28.1)(6, 28.1)(7, 28.1)(17, 28.1)};
        \addlegendentry{synROD $\rightarrow$ ARID}
        
        \addplot[color=my_orange,mark=none]
        coordinates {(3, 28.6)(11,28.6)(13, 28.3)(15,27.5)(17,26.6)(19, 28.1)};
        \addlegendentry{ROD $\rightarrow$ ARID}
        
 \end{axis}
        \end{tikzpicture}%
      }\vspace{-2mm}
\caption{OWR-H results when varying $\varepsilon$ and $b$.}
\label{fig:ablation}
    \end{minipage}
\vspace{-5mm}
\end{figure}
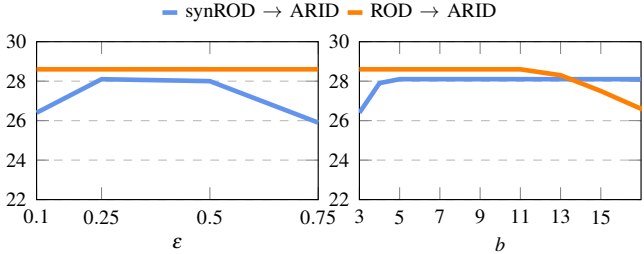

\subsection{Incremental learning performance}
The performance over subsequent incremental steps is an important aspect to consider when comparing incremental learning methods.
We perform this analysis by reporting the scores for all the three metrics (ACC-WR, ACC, OWR-H) in Figure \ref{fig:incremental_trend} on the synROD $\to$ ARID domain shift. In this case, by looking at Table \ref{tab:sota}, we identify the Self Challenging (SC) approach as the DG method providing the higher mean improvement to \OUR's competitors, therefore we select it for the comparison. For all the three metrics we can see that \OUR keeps a large gap of performance over the others for all the incremental steps. Despite the natural decrease in accuracy after the first task, mainly in the ACC-WR metric, in the subsequent tasks \OUR is able to maintain a quite stable ACC and OWR-H performance showing a great ability to balance the accuracy on known and unknown samples. 
We remark that \OUR exploits a very simple replay strategy with a 
fixed-size buffer containing randomly chosen samples of old tasks, thus the results are not the consequence of a sophisticated incremental technique.

\subsection{Sensitivity Analysis} \label{sec:ablation}
In Fig. \ref{fig:ablation} we evaluate the robustness of \OUR when changing the values for its two hyperparameters $\varepsilon$ and $b$. They have different influence on the model performance, contingent on the value of $\lambda$, which in turns depends on the statistics of the training dataset. We consider two different shifts representing two possible extreme cases: for synROD $\rightarrow$ ARID we have $\lambda \approx 1 + \varepsilon$ (a situation similar to Fig. \ref{fig:sphere} left), while for ROD $\to$ ARID we have $\lambda \gg 1$ (Fig. \ref{fig:sphere} right). The value of $\varepsilon$ controls the minimum margin between known class clusters imposed through the stop training criterion of Eq. (\ref{eq:stop_criterion}). A larger $\varepsilon$ will push the training towards a larger margin by increasing clusters compactness and separation. While this is desirable, a too high value may lead to overfitting and exceptionally long times of training in order to meet the stop training condition. The value of $\varepsilon$ does not influence the performance on ROD $\to$ ARID as for this shift the margin is naturally quite high. The second hyperparameter $b$ tunes our known-unknown separation threshold $\tau$ (see Eq. (\ref{eq:threshold})). It allows to find a good balance between known and unknown accuracy for both cases. It can be noticed from the plots
that the results obtained by \OUR are stable and high ($\geq26\%$) for a reasonable range of values, always outperforming the best competitor (e.g. for synROD $\to$ ARID, NNO + SC obtains 16.9, for ROD $\to$ ARID, DeepNNO + RR obtains 25.8).

%\bigskip
\section{CONCLUSIONS}
In this work, we proposed the first approach able to tackle all the challenges of the CD-OWR setting at once. We demonstrated how a simple contrastive learning-based approach represents a very powerful solution for the task. We further introduced self-paced strategies to define both an appropriate stopping criterion and a good threshold for known-unknown separation, obtaining a model able to reach state-of-the-art results. Considering the importance of this topic we believe that our work can pave the way for future investigations.

\vspace{2mm}\noindent\textbf{Acknowledgements}. Computational resources were provided by IIT (HPC infrastructure) and CINECA through the IsC94 Tr-OSDG award under the ISCRA initiative. We also acknowledge the support of the European H2020 Elise project (\url{www.elise-ai.eu}).
We thank Dario Fontanel for the useful discussions.

\bibliographystyle{IEEEtran}
\bibliography{ebib}

\end{document}